\documentclass[12pt,a4paper]{article}

\title{Primitive Part-Of-Speech Tagging using Word Length and Sentential
Structure}
\author{Simon Cozens\footnote{$<simon.cozens@ox.ac.uk>$} \\
Oriental Institute, \\
University Of Oxford,\\
Pusey Lane, Oxford. UK \\}
\date{July 1998}

\begin{document}
\maketitle

\begin{abstract}
It has been argued that, when learning a first language, babies use a
series of small clues to aid recognition and comprehension, and that one
of these clues is word length. 
In this paper we present a statistical part of speech tagger which trains
itself solely on the number of letters in each word in a sentence. 
\end{abstract}

\newcounter{example}
\section{Introduction}

In the modular model of part-of-speech tagging, a number of factors are used in parallel to assist determination. This method has the advantage that it is not reliant on any particular language model, but can gather probabilistic data from a number of very different tests on a text. In this paper we present one such possible module.

While it seems a primitive idea, the number of letters in each word in a sentence is a surprisingly good indicator of part-of-speech patterns. We implemented a tagger which made use of word-length data alone, and reported results of 33\%, which should help to increase the accuracy of a modular part-of-speech parser to near human levels.

\section{Implementation}
The tagger was implemented in several stages:\\

{\bf Corpus preparation}, in which the inital SUSANNE corpus was
reformatted for managability.\\

{\bf Building the knowledge-base} where the tagger was trained by keying
groups of word length information to sentential and intrasentential
part-of-speech structures using the modified corpus.\\

{\bf Implementing the tagger} proper, by searching the knowledge base for
keys with similar word length structures and, by weighting each result by
structure length, combining the results to form the putative tagging of
the test sentence.

\subsection{Corpus Preparation}

The corpus used in the experiments with the tagger was Release 4 of the
SUSANNE corpus \cite{SUSANNE} which contains a 130,000 word subset of the
Brown corpus \cite{Brown} tagged and marked up for part-of-speech and
structure parsing information. \\

The knowledge-base of our statistical parser was built up from the
following Brown genre categories:
\begin{itemize}
\item{A	- press reportage}
\item{G	- belles lettres, biography, memoirs}
\item{J - learned (mainly scientific and technical) writing}
\end{itemize}

The rest of the corpus, that of category N, "adventure and Western
fiction" made up the sentences for testing the tagger. \\

The corpus was prepared in the following stages: \\

{\bf Conversion of SGML markup to ASCII Text.} Since SGML markup affects
the number of characters in each word, the markup was replaced by the
equivalent character in the ASCII characterset. \\

{\bf Reformatting of corpus structure.} To expediate the building of the
training corpus, the structure of the SUSANNE corpus was modified. The
reference number and structure parse information was ignored, and the file
reformatted to a colon-delimited flatfile for ease of manipulation. Other
structural information was also removed at this stage. \\

{\bf Simplification of tagset.} As the tagset of the SUSANNE corpus is
quite comprehensive and carries morphological and inflectional information
and the aim of our tagger was to model primitive word-type resolution, it
was decided to group the initial tagset of 353 distinct tags into a more
managable 15 basic word-types. These reflected more accurately the basic
level of word types being modelled and, by narrowing the variation in the
tagset, would hopefully lead to a higher success ratio. \\

{\bf Adding a word-length count.} To further speed up the learning and
keying process, a field representing the length of each word was added to
each record. It is worth noting that the words themselves could have been
removed from the training corpus at this point, but were left in for
reference. \\

For example, the first 5 lines of the SUSANNE corpus in its intial state
look like this: 
\begin{verbatim}
A01:0010a       -       YB      <minbrk>        -       [Oh.Oh]
A01:0010b       -       AT      The     the     [O[S[Nns:s.
A01:0010c       -       NP1s    Fulton  Fulton  [Nns.
A01:0010d       -       NNL1cb  County  county  .Nns]
A01:0010e       -       JJ      Grand   grand   .
\end{verbatim}

The first 5 lines of the corpus used to train the tagger, on the other
hand, look like this:
\begin{verbatim}
Det :3:"The"
N :6:"Fulton"
N :6:"County"
Adj :5:"Grand"
N :4:"Jury"
\end{verbatim}

\subsection{Building the Knowledge-Base}

The knowledge-base was then built by running a Perl program over the
training corpus which, for each word, first keyed the number of letters in
that word to the relevant part of speech, and then expanded outwards to
words preceding and following, keying the number of letters of each word
of the group to the parts of speech until the whole sentence had been
scanned, and then moved onto the next word. \\
For example, the first 5 lines shown above would be keyed as follows:
\begin{verbatim}
3 -> Det                              (Start from "the")
3:6 -> Det:N                          ("the Fulton")
3:6:6 -> Det:N:N
3:6:6:5 -> Det:N:N:Adj
3:6:6:5:4 -> Det:N:N:Adj:N
...
6 -> N                               (Start from "Fulton")
3:6:6 -> Det:N:N                     ("the Fulton County")
...
6 -> N                               (Start from "County")
6:6:5 -> Det:N:Adj
...
\end{verbatim}
\subsection{Implementing the Tagger}

In the case of simple, single-pass tagging, the actual process of tagging
takes place according to this algorithm: \\

\begin{itemize}
\item{Starting at the first word, look to see if the word lengths
describing the whole sentence can be found in the knowledge-base, and
extract the part of speech for the first word.}
\item{Weight the matches at this stage highly.}
\item{See if the whole sentence excluding the first/last {\it n} words can
be found, again extracting the relevant part of speech.}
\item{Weight these matches less highly.}
\item{Increase {\it n}.}
\item{Go back to step 3 until only the word in question is being examined.
Matches at this level are weighted very low.}
\item{Select the most likely match.}
\item{Move to the next word.}
\end{itemize}

In the tests used, the part of speech score was given as :
\begin{equation} 
score[POS] = \sum^{x=POS}{2^{match length}}
\end{equation}

That is, $2^{match length}$ for each time the part of speech is matched in
the knowledge base.

While not tested at present, it is quite possible that there will exist a
multiple-pass method, whereby once the tag of a word is found by the above
method, when determining the tag of the next word in the sentence, the
tagger looks for keys in the knowledge base describing both the word
length structure and the sentence structure formed by tags already
determined. This method can be used many times to revise the intial
guesses in an almost self-training manner.

\section{Results}

32,777 words were tested from part N of the SUSANNE corpus. 
Using the single-pass methods outlined above, 11,118 were corrected identified
by the tagger - a success rate of 33.92{\%}. While this is not particularly
brilliant in itself, it must be noted that this method should be considered supplemental to conventional part-of-speech determination methods. Using the primitive word-length tagger in conjuction with these, it would not seen unreasonable to estimate that performance could be increased by up to 5{\%}. This would be enough to give human-like abilities to the already existing methods.

\section{Evaluation}

We have presented a very simple and primitive method of part-of-speech tagging using statistical methods alone, and taking into account only a very small part of the data contained within a textual source. This has, however, produced strikingly good results for the scale of the process. 

When combined with conventional part-of-speech determination methods, and other `modular' primitive methods, (morpheme isolation, for example) we would expect to see extremely high rates of success in part-of-speech tagging.

However, we must not forget that the initial corpus was very large, (100,000 words) and produced over 200 megabytes of knowledge-base\footnote{Efficient data handling meant that the data was keyed and indexed; the program never needed to search through the whole knowledge-base, and so completed the 32,000 word test in under 24 hours.}, and therefore there was a high probability that exactly the same phrases turned up in both training and test corpora. It also relied on the feature of English (and most Romance languages) that certain parts of speech are, in the majority of cases, considerably shorter or considerably longer than others. This method alone would obviously not be appropriate in the case of languages such as Japanese.

Nevertheless, this primitive method could have applications as a modular extension to English-language parts of speech determiners, as well as going some way to explaining the way in which the human brain understands and learns language.


\begin{thebibliography}{1}

\bibitem{Brown}
W.~N. Francis and H.~Cucera.
\newblock {\em A Manual of Information to Accompany a Standard Corpus of
  Present-Day Edited American English, for Use with Digital Computers}.
\newblock Department of Linguistics, Brown University, 1989.

\bibitem{SUSANNE}
Geoffrey Sampson.
\newblock {\em English for the Computer : the SUSANNE corpus and analytic
  stream}.
\newblock Clarendon Press, 1995.

\end{thebibliography}
\end{document}